\documentclass[runningheads]{llncs}

\usepackage[utf8]{inputenc}
\usepackage{polski}
\usepackage[english]{babel}

\usepackage{amsmath, amssymb}
\usepackage{graphicx}
\usepackage{float}  % for H dock
\usepackage{textcomp}  % for tilde sign
\usepackage{booktabs}
\usepackage[hidelinks]{hyperref}

\usepackage[linesnumbered,lined,boxed]{algorithm2e}
\usepackage[noend]{algpseudocode}
\usepackage{subcaption}  % for side-by-side figures
\usepackage{todonotes}

\DeclareMathAlphabet{\mathcal}{OMS}{cmsy}{m}{n}

\DeclareMathOperator{\EX}{\mathbb{E}}% expected value

\SetCommentSty{mycommfont}
\SetEndCharOfAlgoLine{}

\makeatletter
\def\BState{\State\hskip-\ALG@thistlm}
\makeatother

\title{Semi-supervised learning \\with Bidirectional GANs}
\author{Maciej Zamorski \and Maciej Zięba}
\institute{Faculty of Computer Science and Management\\Wrocław University of Science and Technology\\
\email{maciej.zamorski@pwr.edu.pl, maciej.zieba@pwr.edu.pl}
\and
Tooploox Ltd.
}

\begin{document}
\mainmatter 
\selectlanguage{english}
\maketitle

\begin{abstract}
In this work we introduce a novel approach to train Bidirectional Generative Adversarial Model (BiGAN) in a semi-supervised manner. The presented method utilizes triplet loss function as an additional component of the objective function used to train discriminative data representation in the latent space of the BiGAN model. This representation can be further used as a seed for generating artificial images, but also as a good feature embedding for classification and image retrieval tasks. We evaluate the quality of the proposed method in the two mentioned challenging tasks using two benchmark datasets: CIFAR10 and SVHN.
\keywords{Generative models \and Triplet learning \and Generative Adversarial Networks \and Image retrieval}
\end{abstract}

\pagenumbering{arabic}
\section{Introduction}
One of the most common and important tasks of machine learning is building generative models, that can capture and learn a wide variety of data distributions. Recent developments in generative modeling concentrate around two major areas of research: variational autoencoders (VAE) \cite{kingma2013auto} that aim at capturing latent representations of data, while simultaneously keeping it restricted to known distribution (e.g., normal distribution) and generative adversarial networks (GANs) \cite{goodfellow2014generative,radford2015unsupervised} with grounds in game theory, having strong emphasis on creating realistic samples from underlying distributions. 

These kind of models are not only known from generating data from the distribution represented by data examples but are also used to train informative and discriminative feature embeddings. It can be obtained either only with unsupervised data by using good discriminative properties of the GAN's discriminator achieved during adversarial training \cite{radford2015unsupervised,zieba2018bingan} or using some subset of labeled data and incorporating semi-supervised mechanisms during training the generative model \cite{salimans2016improved,zieba2017training}.  

In this work we concentrate on obtaining better feature representation for image data using semi-supervised learning with a model based on Bidirectional Generative Adversarial Networks (BiGANs) \cite{donahue2016adversarial} / Adversarially Learned Inference (ALI) \cite{dumoulin2016adversarially}. In order to incorporate semi-supervised data into training procedure, we propose to enrich the primary training objective with an additional triplet loss term \cite{hoffer2015deep} that operates on the labeled examples.

Our approach is inspired by the work \cite{zieba2017training} where the triplet loss was used to increase the quality of features representation in the discriminator. Contrary to this approach, we make use of an additional model in BiGAN architecture - encoder and aim at increasing the quality of feature representation in the coding space, that is further used by a generator to create artificial samples. Practically, it means that the feature representation can be used not only for classification and retrieval purposes but also for generating artificial images similar to existing. 

The contribution of the paper is twofold. We introduce a new GAN training procedure for learning latent representations that extends the models presented in \cite{donahue2016adversarial,dumoulin2016adversarially} and inspired by \cite{zieba2017training} for semi-supervised learning. We show that Triplet BiGAN will result in superior scores in classification and image retrieval tasks.

This work is organized as follows. In Sec. \ref{sec:rw} we present the basic concepts related to GAN models and triplet learning. In Sec. \ref{sec:sstbigan} we describe our approach - Triplet BiGAN model. In Sec. \ref{sec:exp} we provide the results obtained by Triplet BiGAN on two challenging tasks: image classification and image retrieval. This work is summarized in Sec. \ref{sec:summary}.

\section{Related works}
\label{sec:rw}

\subsection{Generative Adversarial Networks}
Since their inception, Generative Adversarial Networks (GANs) \cite{goodfellow2014generative} have become one of the most popular models in a field of generative computer vision. Their main advantages come from their straightforward architecture and ability to produce state-of-the-art results. Studies performed in recent years propose many performance, stability and usage improvements to the original version, with Deep Convolutional GAN (DCGAN) \cite{radford2015unsupervised} and Improved GAN \cite{salimans2016improved} being used most often as architectural baselines in pure image generation learning tasks. 

The main idea of GAN is based on game theory and assumes training of two competing networks: generator $G(\mathbf{z})$ and discriminator $D(\mathbf{x})$. The goal of GANs is to train generator $G$ to sample from the data distribution $p_{data}(\mathbf{x})$ by transforming the vector of noise $\mathbf{z}$ to the data space. The discriminator $D$ is trained to distinguish the samples generated by $G$ from the samples from $p_{data}(\mathbf{x})$. The training problem formulation is as follows:

\begin{equation}\label{eq:v_ge}
\min_G \max_D V(G, D) = \EX_{x \sim p_{data}}[\log D(x)] + \EX_{z \sim p_{z}}[\log(1 - D(G(z)))]
\end{equation}
where: $p_{data}$ - true data distribution, $p_z$ - prior to the data space.

The model is usually trained with the gradient-based approaches by taking minibatch of fake images generated by transforming random vectors sampled from $p_{\mathbf{z}}(\mathbf{z})$ via the generator and minibatch of data samples from $p_{data}(\mathbf{x})$. They are used to maximize $V(D,G)$ with respect to parameters of $D$ by assuming a constant $G$, and then minimizing $V(D,G)$ with respect to parameters of $G$ by assuming a constant $D$.

\subsection{Bidirectional Generative Adversarial Networks}

BiGAN model, presented in \cite{donahue2016adversarial,dumoulin2016adversarially} extends the original GAN model by an additional encoder module $E(\mathbf{x})$, that maps the examples from data space $\mathbf{x}$ to the latent space $\mathbf{z}$. By incorporating the encoder into the GAN architecture, we can code examples in the same space that is used as a seed for generating artificial samples. 

The objective function that is used train the BiGAN model can be defined in the following manner:

\begin{equation} \label{eq:v_gde}
V(G, D, E) = \EX_{x \sim p_x}[\log D(x, E(x))] + \EX_{z \sim p_z}[\log (1 - D(G(z), z))].
\end{equation}

The adversarial paradigm applied to train the model BiGAN is analogous as for GAN model. The goal of training is to solve the min-max problem stated below:

\begin{equation}
\min_{G, E} \max_D V(G, D, E).
\end{equation}

Practically, the model is trained in alternating procedure, where the parameters of discriminator $D$ are updated by optimizing the following loss function:

\begin{equation} \label{eq:bigan_discriminator_loss}
L_D = \EX_{x \sim p_x}[\log D(x, E(x))] + \EX_{z \sim p_z}[\log (1 - D(G(z), z))],
\end{equation}
and the parameters of generator $G$ and encoder $E$ are jointly trained by optimizing the following loss:
\begin{equation} \label{eq:bigan_encgen_loss}
L_{EG} = \EX_{z \sim p_z}[\log D(G(z), z)] + \EX_{x \sim p_x}[\log (1 - D(x, E(x)))].
\end{equation}
Since (as authors showed in \cite{donahue2016adversarial}) the encoder, in order to be an optimal one, learns to invert the examples from true data distribution, the same loss can be applied to the encoder and the generator parameters.

Experiments show that the encoder despite learning in a pure unsupervised way was able to embed meaningful features, which later show during reconstruction. The inclusion of additional module raises a question about the quality of this feature representation for classification and image retrieval tasks. The approach of combining objectives seems promising, as the encoder module is explicitly trained for feature embedding, as opposed to the discriminator, which main task is to categorize samples into real and fake.

\subsection{Triplet Networks}
Triplet networks \cite{hoffer2015deep} are among the most used methods for deep metric learning \cite{kumar2016learning,yao2016deep,zhuang2016fast}. Triplet networks consist of three instances of the same neural network, that share parameters among themselves. During training, the triplet model $T$ receives three examples from the training data: the reference sample $x^q$, the positive sample (the sample that is in some way similar to the reference sample, f.e. it belongs to the same class) $x^+$ and the negative sample (that is dissimilar to the reference sample) $x^-$. The goal is to train the triplet network $T$ in such a way, that the distance $d^-$ between encoded query example $T(x^q)$ to the encoded negative example $T(x^-)$ is greater than the distance $d^+$ form the encoded query example $T(x^q)$ to the encoded positive example $T(x^+)$. In general case, this distances are computed as a L2-norm between feature vectors, i.e. : $d^- = \| T(x) - T(x^-) \|_2 $ and $d^+ = \| T(x) - T(x^+) \|_2$. 

During the training the triplet model makes use of the probability $p_T$ that the distance of the query example to the negative example is greater than its distance to the positive one which can be defined in the following way:

\begin{equation} \label{eq:pb_triplet}
p_T = \frac{\exp(d^-)}{\exp(d^+) + \exp(d^-)}
\end{equation}

We formulate the objective function for a single triplet $(x, x^+, x^-)$ in a following manner \cite{zieba2017training}:

\begin{equation} \label{eq:pb_triplet_loss_1}
L_T = -\log (p_{T(x, x^+, x^-)})
\end{equation}

The parameters of the model $T$ are updated according to the gradient-based approach that is used to optimize the objective function $L_T$ by utilizing the minibatches of triplets $(x, x^+, x^-)$ selected from data. Usually, the procedure of triplet selection is performed randomly (assuming that $x$, $x^+$ are closer than $x$, $x^-$) but there are some other approaches that speed-up the training process. The most popular is to construct the triplets for training taking under consideration the hardest negative samples $x^-$, which are the closest to the currently selected reference sample $x$.  

\begin{figure}[ht]
\centering
\includegraphics[width=.7\textwidth]{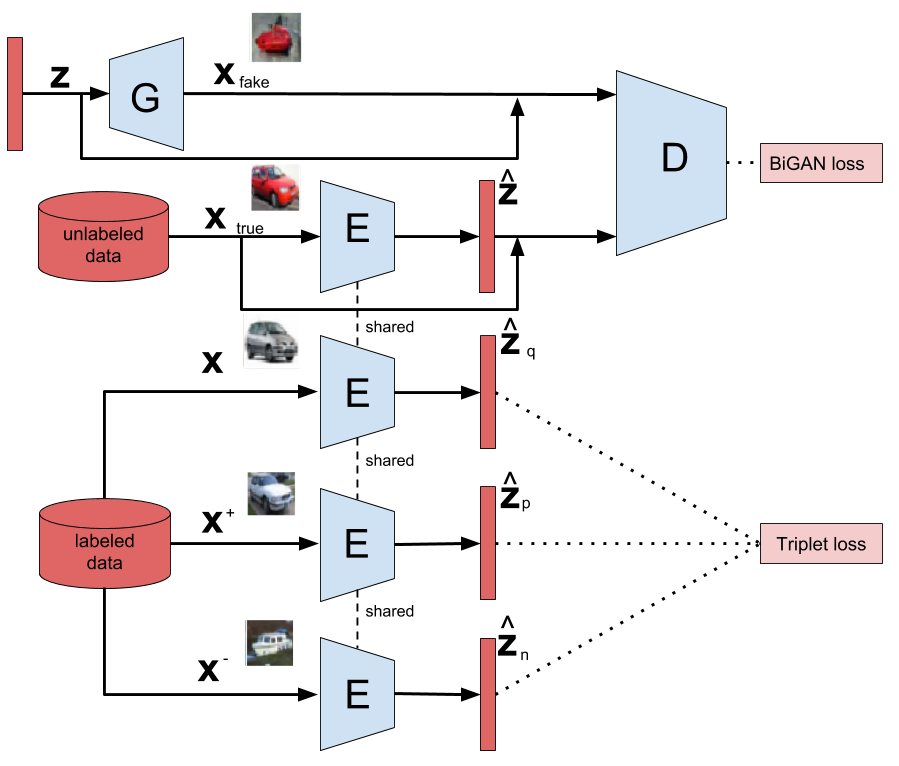}
\caption[model]{The scheme presents the architecture of Triplet BiGAN model. We modify the BiGAN architecture by incorporating the encoding model in additional triplet task. We use the labeled part of the training data to construct triplets $(x, x^+, x^-)$, where we assume that $x$ and $x^+$ are from the same class and $x$ and $x^-$ are from different classes. Each of component of the triplet $(x, x^+, x^-)$ is passed thru the encoding module to obtain the corresponding coding vectors $(z, z^+, z^-)$ that are further used to construct the triplet loss. The remaining part of the model operates on the unsupervised data as in basic BiGAN model. }
\label{fig:triplet_bigan_arch}
\end{figure}

\section{Triplet BiGANs}
\label{sec:sstbigan}

In this work we introduce Triplet BiGAN model that combines the benefits of using BiGAN in terms of learning interesting representations in latent space and the superior power of the triplet model that trains well using supervised data. The core idea of our approach is to incorporate the encoder model of BiGAN to act as triplet network on the labeled part of training data (see fig. \ref{fig:triplet_bigan_arch}). 

In terms of training the Triplet BiGAN we simply modify the $L_{EG}$ (see eq. (\ref{eq:bigan_encgen_loss})) criterion by incorporating an additional triplet term:

\begin{equation} \label{eq:triplet_bigan_encgen_loss}
L_{TEG} = L_{EG} + \lambda \cdot \EX_{(x, x^+, x^-) \sim p_{triplet}}[ -\log (p_{T(x, x^+, x^-)})],
\end{equation}
where $p_{T(x, x^+, x^-)}$ is triplet loss defined by eq. (\ref{eq:pb_triplet_loss_1}), $\lambda$ is a hyperparameter that represents the impact of the triplet loss on the global criterion and $p_{triplet}$ is the distribution that generates triplets, where $x$, $x^+$ are from the same class and $x$, $x^-$ are from different classes. 

 Triplet BiGAN model is dedicated to solving semi-supervised problems, where only some portion of labeled data is available. Practically, we do not have access to $p_{triplet}$, therefore we are sampling the triplets $(x, x^+, x^-)$ from some portion of an available labeled dataset, $X_q$.

\begin{algorithm}
$\theta_G$, $\theta_E$, $\theta_D$ $\leftarrow$ \textnormal{initialize network parameters}\;

\While{\textnormal{converging}}{
	\For{$(x, x_q)$ in $(X, X_q)$}{

    	$z \sim \mathcal{N}(0, I)$ \tcp*[f]{Sample feature data from normal distribution}\;
        $\hat{x} \leftarrow G(z)$ \tcp*[f]{Generate images using sampled data}\;
        $\hat{z} \leftarrow E(x)$ \tcp*[f]{Encode images from a minibatch}\;
        \BlankLine
        
        $x_+ \leftarrow$ random\ $x_q' \in X_q$, same class as $x_q$ \;
        $x_- \leftarrow$ random\ $x_q' \in X_q$, different class than $x_q$ \;
        \BlankLine
        
        $d_+ \leftarrow \| E(x_q) - E(x_+) \|_2$ \;
        $d_- \leftarrow \| E(x_q) - E(x_-) \|_2$ \;
        \BlankLine
        
        $L_D \leftarrow \log D(x, \hat{z}) + \log (1 - D(\hat{x}, z))$ \tcp*[f]{GAN loss for $D$}\;
        $L_{EG} \leftarrow \log D(\hat{x}, z) + \log (1 - D(x, \hat{z}))$  \tcp*[f]{GAN loss for $G$, $E$}\;
        $L_T \leftarrow -\log \frac{\exp(d_-)}{\exp(d_+) + \exp(d_-)}$ \tcp*[f]{Triplet loss for $E$} \;
        \BlankLine
        
        $\theta_D \leftarrow \theta_D - \nabla_{\theta_D}L_D$\;
        $\theta_G \leftarrow \theta_G - \nabla_{\theta_G}L_{EG}$\;
        $\theta_E \leftarrow \theta_E - \nabla_{\theta_E}(L_{EG} + L_T)$\;
    }
}
\caption{Training procedure for Triplet BiGAN. }
\label{alg:triplet_bigan}
\end{algorithm}

The training procedure for the model is described in alg. \ref{alg:triplet_bigan}. We assume that $G$, $E$, $D$ are neural networks, that are described by the parameters $\theta_G$, $\theta_E$, $\theta_D$ respectively. For the training procedure we assume, that we have access to unsupervised data $X$ and some portion of supervised data $X_q$. For each training iteration, we randomly sample noise vector $z$ from the normal distribution, pass it thru generator $G$ to obtain the fake sample $\hat{x}$. We select $x$ from unlabeled data $X$ and triplet $(x, x_+, x_-)$ from labeled data $X_q$. Using encoder $E$ we receive the coding vector $z$ corresponding to the sample $x$. Next, we update the parameters $\theta_D$ of discriminator $D$ by optimizing the criterion $L_D$. During the same iteration we update the parameters of generator $G$ and encoder $E$ by optimizing $L_{TEG} = L_{EG} + L_T$. The procedure is repeated until convergence.

In practical implementation, we make use stochastic gradient optimization techniques and perform gradient updates using ADAM method. We also initialize parameters of the Triplet BiGAN by training simple BiGAN without triplet term for given number of epochs without triplet term ($\lambda=0$).

The motivation behind this approach is to increase the discriminative capabilities of the codes obtained from latent space for BiGAN model using some portion of labeled examples involved in triplet training. As a result, we obtain the encoding model $E$, that is not only capable of coding the data examples for further reconstruction but also can be used as good quality feature embedding for the tasks like image classification or retrieval.

\section{Experiments}
\label{sec:exp}
The goal of the experiments is to evaluate the discriminative properties of the encoder in two challenging tasks: image retrieval and classification. We compare the results with the two reference approaches: triplet network trained only with supervised data and simple BiGAN model, where the latent representation of encoder is used for evaluation.     

\subsubsection{Datasets}
The model was trained on two datasets: Street View House Numbers (SVHN) and CIFAR10. In each dataset, 50 last examples of each class were used as a validation set and were not used for training the models. During training only selected portion of training set have assigned labels. The next subsection presents results obtained when using only 100, 200, 300, 400 or 500 labeled examples per class. For testing purposes, we trained classifier only on the images from the training split, that were given a label for triplet training.

\subsubsection{Metrics}
Retrieval evaluation was done with accuracy and mean average precision (mAP). For classification, 9-nearest neighbors classifier was used with weighted by the distance-based importance of neighbors. Mean average precision was calculated at length of encoded data. Cluster visualization was performed by applying t-SNE\cite{maaten2008visualizing} with Euclidean metric, perplexity 30 and Barnes-Hut approximation for 1000 iterations. 

\subsubsection{Architecture}
The architectures of discriminator, encoder, and generator were as presented in \cite{dumoulin2016adversarially}. 

The encoder network $E$ is a 7-layer convolutional neural network, that learns the mapping from image space $X$ to feature space $z$. After each convolutional layer (excluding the last), a batch normalization is performed, and the output is passed through a leaky relu activation function. After penultimate convolutional block (meaning convolutional layer with normalization and activation function) a reparametrization trick \cite{kingma2013auto} is performed.

The generator network $G$ is a neural network with seven convolution transposition layers. After each layer (except the last) a batch normalization is performed, and the output is passed through a leaky relu activation function. After the last convolution-transposition layer, we squash the features to $(0, 1)$ range with the sigmoid function.

The discriminator part $D$ consists of three neural networks -- $D_x$ discriminates in the image space, $D_z$ discriminates in the encoding space. Both of them map their inputs into a discriminative latent space and each of them returns a same-size vector. Third network $D_{xz}$ takes concatenation of said vectors as an input and returns a decision, whether an input tuple (image, encoding) comes from encoding or generative part of the Triplet BiGAN network. Image discriminator $D_x$ is made of five convolution layers with Leaky Relu nonlinearity after each of them. Encoding discriminator is represented as two convolution layers with Leaky Relu nonlinearity after each of them and Joint discriminator $D_{xz}$ is another three convolutional layers with Leaky Relu between them and the sigmoid nonlinearity at the end.

\subsection{Results}
%This section contains results of experiments described in the previous chapter. 
\subsubsection{Classification}
For assessing classification accuracy quality, the experiments were done to test the influence of feature vector size and images per class taken for semi-supervised learning. For each of the model, the experiments were done, when the feature vector consisted of either 16, 32, 64 or 128 (256 for SVHN) variables and 500 labeled images per class were taken. On the other hand, using feature vector size of 64, the experiments measured the impact of a number of labeled examples available during training, with possible values being 100, 200, 400 and 500 (only for Cifar10). The experiments were conducted on Cifar10 and SVHN datasets.

\begin{table}[!htb]
\centering

\caption{Classification results on CIFAR10 dataset for different sizes of encoder feature vector using 500 labeled examples per class in the training set. $m$ - vector size. Only labeled samples from training set used.}
\label{classification_cifar_vector}
\begin{tabular}{|c|r|r|r|r|}
\hline
\textbf{Model}                  & \multicolumn{1}{c|}{\textbf{m=16}} & \multicolumn{1}{c|}{\textbf{m=32}} & \multicolumn{1}{c|}{\textbf{m=64}} & \multicolumn{1}{c|}{\textbf{m=128}} \\ \hline
Triplet                             & 44.42                              & 45.56                             & 52.32                              & 46.15                              \\  \hline
BiGAN                               & 41.30                             & 43.81                             & 48.50                            & 49.13                            \\ \hline
\multicolumn{1}{|l|}{Triplet BiGAN} & 61.08                              & 62.40                              & 63.14                             & 53.92                              \\ \hline
\end{tabular}
\end{table}

\begin{table}[!htb]
\centering

\caption{Classification results on SVHN dataset for different sizes of encoder feature vector using 200 labeled examples per class in the training set. $m$ - vector size. Only labeled samples from training set used.}
\label{classification_svhn_vector}
\begin{tabular}{|c|r|r|r|r|}
\hline
\textbf{Model}                  & \multicolumn{1}{c|}{\textbf{m=16}} & \multicolumn{1}{c|}{\textbf{m=32}} & \multicolumn{1}{c|}{\textbf{m=64}} & \multicolumn{1}{c|}{\textbf{m=256}} \\ \hline
Triplet                             & 66.12                              & 69.96                              & 71.11                              & 71.54                               \\ \hline
BiGAN                               & 44.65                              & 53.80                              & 62.35                               & 17.48                               \\ \hline
\multicolumn{1}{|l|}{Triplet BiGAN} & 71.43                              & 75.65                              & 79.12                               & 78.86                               \\ \hline
\end{tabular}
\end{table}

\begin{table}[!htb]
\centering

\caption{Classification results on CIFAR10 dataset for different portions of labeled samples per class and feature vector size $m=64$. $n$ - number of labeled samples per class. Only labeled samples from training set used.}
\label{classification_cifar_class}
\begin{tabular}{|c|r|r|r|r|}
\hline
\textbf{Model}                  & \multicolumn{1}{c|}{\textbf{n=100}} & \multicolumn{1}{c|}{\textbf{n=200}} & \multicolumn{1}{c|}{\textbf{n=400}} & \multicolumn{1}{c|}{\textbf{n=500}} \\ \hline
Triplet                             & 28.80                             & 34.81                           & 40.12                            & 52.32   \\
\hline
BiGAN                               & 42.73                             & 45.06                             & 47.67                              & 41.30                               \\ \hline
\multicolumn{1}{|l|}{Triplet BiGAN} & 53.15                             & 52.02                              & 57.45                              & 63.14                              \\ \hline
\end{tabular}
\vspace{-4mm}
\end{table}

\begin{table}[!htb]
\centering
\caption{Classification results on SVHN dataset for different portions of labeled samples per class and feature vector size $m=64$. $m$ - vector size. Only labeled samples from training set used.}
\label{classification_svhn_class}
\begin{tabular}{|c|r|r|r|}
\hline
\textbf{Model}                  & \multicolumn{1}{c|}{\textbf{n=100}} & \multicolumn{1}{c|}{\textbf{n=200}} & \multicolumn{1}{c|}{\textbf{n=400}} \\ \hline
Triplet                             & 66.48                              & 71.11                              & 73.76                                                          \\ \hline
BiGAN                               & 61.16                              & 62.35                              & 58.94                                                         \\ \hline
\multicolumn{1}{|l|}{Triplet BiGAN} & 72.40                              & 79.12                              & 82.21
\\ \hline
\end{tabular}
\end{table}

\subsubsection{Retrieval}

For assessing image retrieval quality, the experiments were made to test an influence of feature vector size and images per class taken for semi-supervised learning. For each sample in the testing data, an algorithm sorts the images from the training dataset from closest to most further. Distances are calculated basing on Euclidean distances between images' feature vectors to check if samples that are close to each other in data space (images belong to the same class) are close to each other in feature space (their representation vectors are similar). In ideal type situation (mAP = 1), all of the relevant training images would be put first and only then training images that belong to the same class. With 10 classes in each dataset mAP = 0.1 may be considered random ordering, as it roughly means that on average, only every tenth image was of the same class as the test image.

Results presented in tables below indicate the increased average precision of image retrieval when using Triplet BiGAN method as opposed to using only labeled examples by 0.05 - 0.15 in all, but one experiment.

\begin{table}[!htb]
\centering

\caption{Image retrieval results on CIFAR10 dataset for different sizes of encoder feature vector using 500 labeled examples per class in the training set. $m$ - vector size. Only labeled samples from training set used.}
\label{retrieval_cifar_vector}
\begin{tabular}{|c|r|r|r|r|}
\hline
\textbf{Model}                  & \multicolumn{1}{c|}{\textbf{m=16}} & \multicolumn{1}{c|}{\textbf{m=32}} & \multicolumn{1}{c|}{\textbf{m=64}} & \multicolumn{1}{c|}{\textbf{m=128}} \\ \hline
Triplet                             & 0.4993                              & 0.5134                              & 0.5235                              & 0.5197                               \\  \hline
BiGAN                               & 0.1458                              & 0.1433                              & 0.1634                              & 0.1620                               \\ \hline
\multicolumn{1}{|l|}{Triplet BiGAN} & 0.6292                              & 0.6457                              & 0.6528                              & 0.3748                               \\ \hline
\end{tabular}
\end{table}

\begin{table}[!htb]
\centering
\caption{Image retrieval results on SVHN dataset for different sizes of encoder feature vector using 200 labeled examples per class in the training set. $m$ - vector size. Only labeled samples from training set used.}
\label{retrieval_svhn_vector}
\begin{tabular}{|c|r|r|r|r|}
\hline
\textbf{Model}                  & \multicolumn{1}{c|}{\textbf{m=16}} & \multicolumn{1}{c|}{\textbf{m=32}} & \multicolumn{1}{c|}{\textbf{m=64}} & \multicolumn{1}{c|}{\textbf{m=256}} \\ \hline
Triplet                             & 0.6855                              & 0.7224                              & 0.7474                              & 0.6989                               \\ \hline
BiGAN                               & 0.1492                              & 0.1582                              & 0.1748                              & 0.1633                               \\ \hline
\multicolumn{1}{|l|}{Triplet BiGAN} & 0.7201                              & 0.7633                              & 0.8002                              & 0.7931                               \\ \hline
\end{tabular}

\end{table}

\begin{table}[!htb]
\centering
\caption{Image retrieval results on CIFAR10 dataset for different portions of labeled samples per class and feature vector size $m=64$. $n$ - number of labeled samples per class. Only labeled samples from training set used.}
\label{retrieval_cifar_class}
\begin{tabular}{|c|r|r|r|r|}
\hline
\textbf{Model}                  & \multicolumn{1}{c|}{\textbf{n=100}} & \multicolumn{1}{c|}{\textbf{n=200}} & \multicolumn{1}{c|}{\textbf{n=400}} & \multicolumn{1}{c|}{\textbf{n=500}} \\ \hline
Triplet                             & 0.3732                              & 0.4027                              & 0.4778                              & 0.5235                               \\ \hline
BiGAN                               & 0.1681                              & 0.1666                              & 0.1645                              & 0.1634                               \\ \hline
\multicolumn{1}{|l|}{Triplet BiGAN} & 0.5628                              & 0.5487                              & 0.5966                              & 0.6528                               \\ \hline
\end{tabular}
\end{table}

\begin{table}[!htb]
\centering
\caption{Image retrieval results on SVHN dataset for different portions of labeled samples per class and feature vector size $m=64$. $n$ - number of labeled samples per class. Only labeled samples from training set used.}
\label{retrieval_svhn_class}
\begin{tabular}{|c|r|r|r|r|}
\hline
\textbf{Model}                  & \multicolumn{1}{c|}{\textbf{n=100}} & \multicolumn{1}{c|}{\textbf{n=200}} & \multicolumn{1}{c|}{\textbf{n=400}} \\ \hline
Triplet                             & 0.6323                              & 0.7474                              & 0.7490
\\ \hline
BiGAN                               & 0.1737                              & 0.1748                              & 0.1655  
\\ \hline
\multicolumn{1}{|l|}{Triplet BiGAN} & 0.7300                              & 0.8002                              & 0.8198  \\  \hline
\end{tabular}
\end{table}

\begin{figure}
\centering
\includegraphics[width=\textwidth]{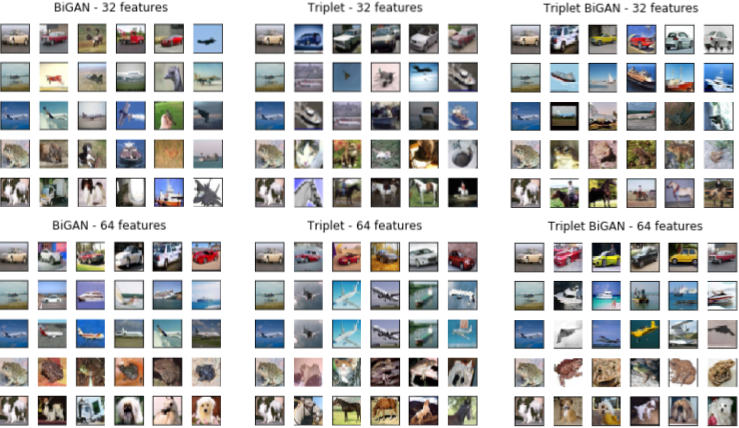}
\caption[Hehe]{Visualization presenting image retrieval results. Left-most column of each section contains five randomly chosen images from the Cifar10 test set. Columns from second-left to right-most present images from the restricted (500 examples per class) training set, from the closest to 5th-closest image.}
\label{fig:retrieval}
\end{figure}

Figure \ref{fig:retrieval} presents a visualization of the closest images from the restricted training set (used only 500 examples per class from the original training set, the same examples that were used in semi-supervised learning). The closeness of the image was decided between each image from the randomly chosen sample of 5 images from the test set and each image from the restricted training set. The distance between images was calculated by encoding each image to feature vector form and calculating Euclidean distance between selected test and training images.

As seen in the visualization, BiGAN model, despite the fact of being trained in a pure unsupervised way, can still embed similar images to similar vectors. However, the closest pictures tend to contain occasional errors, which is not the case with retrieval using triplet models, that tend to contain errors sparingly.

The notable example, showing better results of Triplet BiGAN in comparison to regular Triplet model is the 4th image from the selected test pictures (grey frog). Using 32 and 64 size features vectors Triplet BiGAN was able to retrieve other frog and toad images correctly. The same image caused problems for the original Triplet model, not to mention BiGAN. This shows that additional unsupervised learning of underlying data architecture is indeed beneficial to finding subtle differences in images and can improve the quality of feature embedding.

\subsubsection{Clustering}

\begin{figure}
\centering
\begin{subfigure}{.5\textwidth}
  \centering
\includegraphics[width=.8\textwidth]{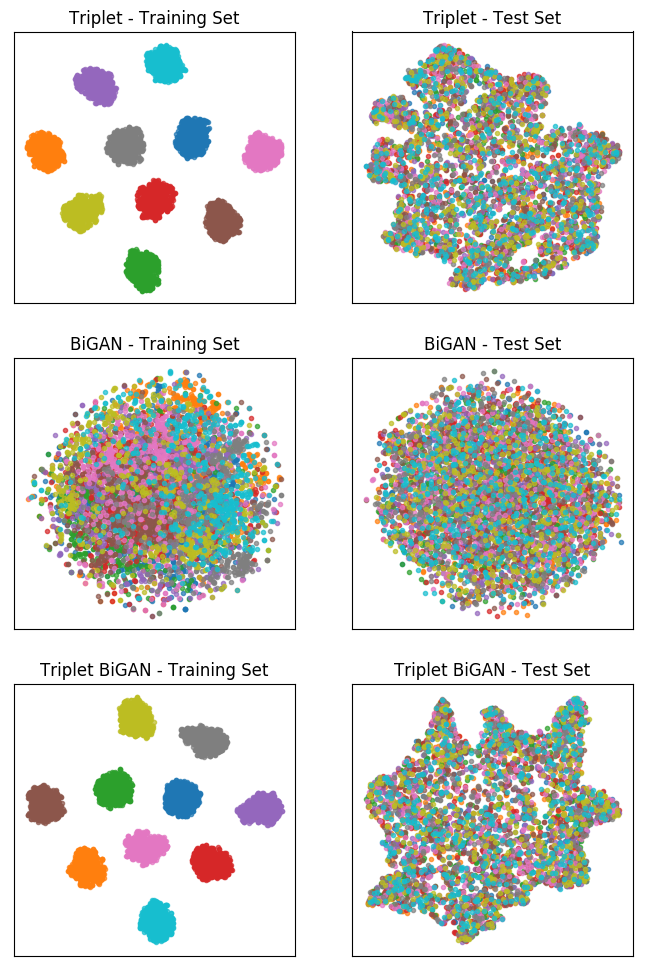}

\end{subfigure}%
\begin{subfigure}{.5\textwidth}
  \centering
\includegraphics[width=.8\textwidth]{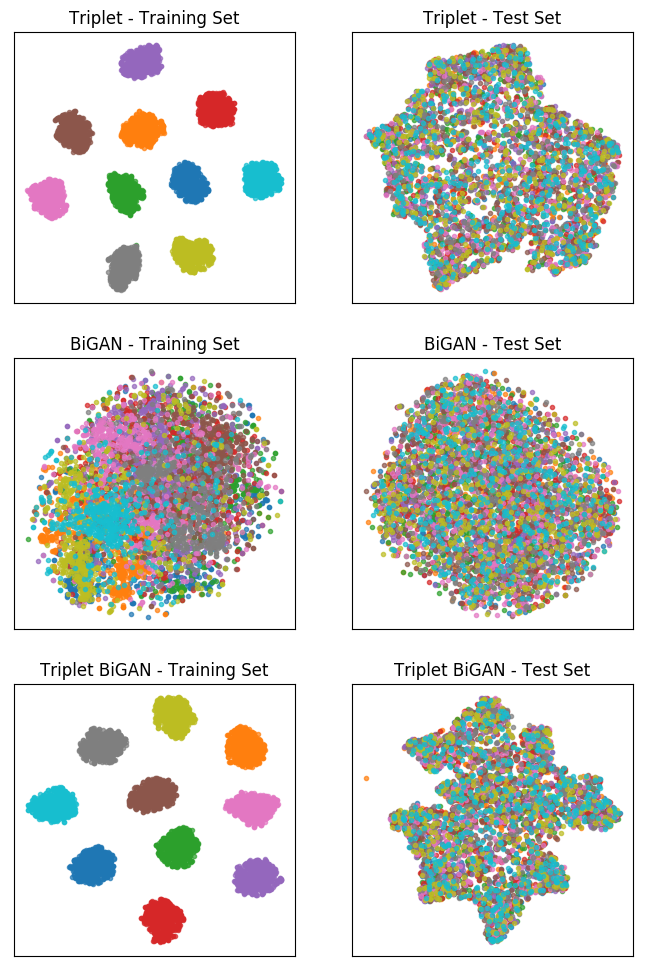}
\end{subfigure}
\caption{a) Clusterization results for models with feature vector size $m = 32$ and trained (Triplet and Triplet BiGAN) using 500 labeled examples per class. \\ b) Clusterization results for models with feature vector size $m = 64$ and trained (Triplet and Triplet BiGAN) using 500 labeled examples per class.}

\end{figure}
Figures below show visualizations of embedding quality of tested models. Each sub-figure present embedding mapped to 2 dimensions using the t-SNE algorithm with one of the three models: Triplet, BiGAN and Triplet BiGAN on the training set and the test set. For Triplet and Triplet BiGAN models, 500 labeled samples per class were used. Two experiments were performed: one with the feature vector of size 32, and one with the feature vector of size 64, as mentioned in figure captions. T-SNE algorithm ran for 1000 epochs using perplexity of 30 and Euclidean metric for distance calculation. In the visualization each class was marked with own color, that was preserved through all sub-figures. 

\subsubsection{Discussion}
In classification and retrieval experiments Triplet BiGAN achieved worse results (tables \ref{classification_cifar_vector}, \ref{classification_cifar_class}, \ref{retrieval_cifar_vector}, \ref{retrieval_cifar_class}) than a Triplet GAN presented in \cite{zieba2017training}. However, we believe than our proposed model has still several advantages in comparison to the reference method. Since in Triplet BiGAN, we perform metric learning on the Encoder (unlike in \cite{zieba2017training} where metric learning is done on the Discriminator features) 

As visualizations suggest, both Triplet and Triplet BiGAN models did not have any problems with learning clusterization on training sets. The output from the t-SNE clearly shows separate group for each class of the samples for this models. This is not the case in BiGAN model. However, while BiGAN was trained without distance-based objective, one can still spot concentration of particular colors. This aligns with observations \cite{dumoulin2016adversarially} that the encoder learns to embed meaningful features into the feature vector, including those, that are somewhat characteristic for specific classes.

Regarding test sets, Triplet and Triplet BiGAN did not generalize to create perfect separations of classes. The models learn to rather bind particular classes into small, homogeneous groups, which are not clearly visible on visualizations but are enough to perform classification using the nearest neighbor algorithm. In the case of BiGAN model the embedding features from the training set do not translate well to the test set, creating a somewhat chaotic collection of points, that is able to generate image retrieval results that are close to random.

\section{Summary}
\label{sec:summary}
This work presents the Triplet BiGAN model that uses joint optimizing criteria: to learn to generate and encode images and to be able to recognize the similarity of given objects. Experiments show that features extracted by an encoder, despite learning only on true data (in opposition to features learned by the discriminator, that learns on real and generated data), may be used as a basis of image classifier, retrieval, grouping or autoencoder model. 

Also included in this work are descriptions of the models that were essential milestones in the field of generative models and distance learning models and an inspiration for creating the presented framework.
\bibliographystyle{splncs04}  
\bibliography{biblography}

\end{document}